\documentclass[11pt,letterpaper]{article}
\usepackage{naaclhlt2015}
\usepackage{times}
\usepackage[skip=0pt]{caption}
\usepackage[table]{xcolor}
\usepackage{color,graphicx}
\usepackage{latexsym,multirow,framed,amsmath,amsfonts,amssymb}
\usepackage{array}
\newcolumntype{L}[1]{>{\raggedright\let\newline\\\arraybackslash\hspace{0pt}}m{#1}}
\newcolumntype{C}[1]{>{\centering\let\newline\\\arraybackslash\hspace{0pt}}m{#1}}
\newcolumntype{R}[1]{>{\raggedleft\let\newline\\\arraybackslash\hspace{0pt}}m{#1}}
\title{Solving Hard Coreference Problems}

\author{Haoruo Peng\thanks{$\;\;$These authors contributed equally to this work.}    $\;$\textnormal{and} Daniel Khashabi$^*$ \textnormal{and} Dan Roth\\
	    University of Illinois, Urbana-Champaign\\
	    Urbana, IL, 61801\\
	    {\tt \{hpeng7,khashab2,danr\}@illinois.edu}}

\date{}

\newif\iflogvar
\logvartrue

\title{ \vspace*{-0.5in}
{{\small \hfill NAACL'15}\\
\vspace*{.25in}} Solving Hard Coreference Problems }

\begin{document}
\maketitle
\begin{abstract}
Coreference resolution is a key problem in natural language understanding that still escapes reliable solutions.
One fundamental difficulty has been that of resolving instances involving pronouns since they often require deep language understanding and use of background knowledge.
In this paper we propose an algorithmic solution that involves a new representation for the knowledge required to address hard coreference problems, along with a constrained optimization framework that uses this knowledge in coreference decision making. Our representation, Predicate Schemas, is instantiated with knowledge acquired in an unsupervised way, and is compiled automatically into constraints that impact the coreference decision.
We present a general coreference resolution system that significantly improves state-of-the-art performance on hard, \textit{Winograd}-style, pronoun resolution cases, while still performing at the state-of-the-art level on standard coreference resolution datasets. 
\end{abstract}

\section{Introduction} \label{sec:intro}
Coreference resolution is one of the most important tasks in \textit{Natural Language Processing} (NLP). Although there is a plethora of works on this task \cite{SoonNgLi01,ng2002identifying,ng2004learning,BengtsonRo08,PMXUZ12,kummerfeld2013error,ChangSaRo13}, it is still deemed an unsolved problem due to 
intricate and ambiguous nature of natural language text. 
%the complicated structure of natural text and a multitude of reasons for mistakes. 
Existing methods perform particularly poorly on 
pronouns, specifically when 
%pronouns where 
gender or plurality information cannot help. In this paper, we aim to improve coreference resolution by 
addressing 
%solving 
these hard problems. Consider the following examples:

\parbox{2.8in}{ 
Ex.1 \textit{[A bird]$_{e_1}$ perched on the [limb]$_{e_2}$ and [it]$_{pro}$ bent.} \\
%\noindent\small
Ex.2 \textit{[Robert]$_{e_1}$ was robbed by [Kevin]$_{e_2}$, and [he]$_{pro}$ is arrested by police.}
}

In both examples, one cannot resolve the pronouns based on only gender or plurality information. Recently, \newcite{rahman2012resolving} gathered a dataset containing 1886 sentences of such challenging pronoun resolution problems (referred to later as the \textit{Winograd} dataset, following Winograd (1972) and Levesque et al. (2011)). As an indication to the difficulty of these instances, we note that a state-of-the-art coreference resolution system \cite{ChangSaRo13} achieves precision of 53.26\% on it. A special purpose classifier \cite{rahman2012resolving} trained on this data set achieves 73.05\%. The key contribution of this paper is a general purpose, state-of-the-art coreference approach which, at the same time, achieves precision of 76.76\% on these hard cases.
%which also achieves precision of 76.76\%. 
 
Addressing these hard coreference problems requires significant amounts of background knowledge, along with an inference paradigm that can make use of it in supporting the coreference decision. 
Specifically, in Ex.1 one needs to know that ``a limb bends" is more likely than ``a bird bends". In Ex.2 one needs to know that the {\em subject} of the verb ``rob" is more likely to be the {\em object} of ``arrest" than the {\em object} of the verb ``rob" is.  The knowledge required is, naturally, centered around the key predicates in the sentence, motivating the central notion proposed in this paper, that of {\em Predicate Schemas}. In this paper, we develop the notion of {\em Predicate Schemas}, instantiate them with automatically acquired knowledge, and show how to compile it into constraints that are used to resolve coreference within a general \textit{Integer Linear Programming} (ILP) driven approach to coreference resolution. Specifically, we study two types of Predicate Schemas that, as we show, cover a large fraction of the challenging cases. 
The first specifies one predicate with its subject and object, thus providing information on the subject and object preferences of a given predicate. The second specifies two predicates with
a semantically shared argument (either subject or
object), thus specifies role preferences of one predicate, among roles of the other. 
We instantiate these schemas by acquiring statistics in an unsupervised way from multiple resources including the Gigaword corpus, Wikipedia, Web Queries and polarity information. 

A lot of recent work has attempted to utilize similar types of resources to improve coreference resolution \cite{rahman2011coreference,RatinovRo12,BansalKl12,rahman2012resolving}. 
The common approach has been to inject knowledge as features. However, these pieces of knowledge provide relatively strong evidence that loses impact in standard training due to sparsity. Instead, we compile our Predicate Schemas knowledge automatically, at inference time, into constraints, and make use of an ILP driven framework \cite{RothYi04} to make decisions. 
Using constraints is also beneficial when the interaction between multiple pronouns is taken into account when making global decisions. Consider the following example:

\parbox{2.8in}{ 
%{\noindent\small
Ex.3 \textit{[Jack]$_{e_1}$ threw the bags of [John]$_{e_2}$ into the water since [he]$_{pro_1}$ mistakenly asked [him]$_{pro_2}$ to carry [his]$_{pro_3}$ bags.}
%}
}

In order to correctly resolve the pronouns in Ex.3, one needs to have the knowledge that ``\textit{he} \underline{asks} \textit{him}'' indicates that \textit{he} and \textit{him} refer to different entities (because they are subject and object of the same predicate; otherwise, \textit{himself} should be used instead of \textit{him}). 
This knowledge, which can be easily represented as constraints during inference, then impacts other pronoun decisions in a global decision with respect to all pronouns: $pro_3$ is likely to be different from $pro_2$, and is likely to refer to $e_2$. This type of inference can be easily represented as a constraint during inference, but hard to inject as a feature.
%This example shows that interdependence of pronouns is often a challenge. Due to the global nature of inference, many works \cite{CWHM07,poon2008joint,BengtsonRo08} advocate global inference for coreference. 

We then incorporate all constraints into a general coreference system \cite{ChangSaRo13} utilizing the mention-pair model \cite{NgCa02,BengtsonRo08,stoyanov2010coreference}. A classifier learns a pairwise metric between mentions, and during inference, we follow the framework proposed in \newcite{CSRRSR11} using ILP.

The main contributions of this paper can be summarized as follows: 
\begin{enumerate}
 \setlength{\itemsep}{0.01pt}
  \setlength{\parskip}{0pt}
  \setlength{\parsep}{0pt}
\item We propose the Predicate Schemas representation and study two specific schemas that are important for coreference. 
\item We show how, in a given context, Predicate Schemas can be automatically compiled into constraints and affect inference. 
%Generate constraints automatically in inference based on acquired knowledge and context.
\item Consequently, we address hard pronoun resolution problems as a standard coreference problem and develop a system\footnote{Available at http://cogcomp.cs.illinois.edu/page/software\newline\_view/Winocoref} which shows significant improvement for hard coreference problems while achieving the same state-of-the-art level of performance on standard coreference problems.
\end{enumerate}

The rest of the paper is organized as follows. We describe our Predicate Schemas in Section \ref{sec:schema} and explain the inference framework and automatic constraint generation in Section \ref{sec:inf}. A summary of our knowledge acquisition steps is given in Section \ref{sec:know}. We report our experimental results and analysis in Section \ref{sec:exp}, and review related work in Section \ref{sec:rel}.

\section{Predicate Schema} \label{sec:schema}
\begin{table*} %[htbp]
\centering
\footnotesize
\begin{tabular}{|c|c|C{12.8cm}|}
\hline
Category & \# & Sentence \\
\hline 
\hline  
\multirow{2}{*}{1} & 1.1 & \textit{[The bird]$_{e_1}$ perched on the [limb]$_{e_2}$ and [it]$_{pro}$ bent.}   \\ 
\cline{2-3} 
& 1.2 & \textit{[The bee]$_{e_1}$ landed on [the flower]$_{e_2}$ because [it]$_{pro}$ had pollen.}  \\ 
\hline
\multirow{2}{*}{2} & 2.1 & \textit{[Bill]$_{e_1}$ was robbed by [John]$_{e_2}$, so the officer arrested [him]$_{pro}$.} \\ 
\cline{2-3}
& 2.2 & \textit{[Jimbo]$_{e_1}$ was afraid of [Bobbert]$_{e_2}$ because [he]$_{pro}$ gets scared around new people. } \\ 
\hline 
\multirow{2}{*}{3}  & 3.1  &  \textit{ [Lakshman]$_{e_1}$ asked [Vivan]$_{e_2}$ to get him some ice cream because [he]$_{pro}$ was hot.} \\
\cline{2-3}
& 3.2 & \textit{ Paula liked [Ness]$_{e_1}$ more than [Pokey]$_{e_2}$ because [he]$_{pro}$ was mean to her. } \\
\hline 
\end{tabular}
\caption{Example sentences for each schema category. The annotated entities and pronouns are hard coreference problems.}
\label{tab:categories:2}
\end{table*}

\begin{table*} %[htbp]
\footnotesize
\centering
\begin{tabular}{|c|c|C{10.1cm}|}
\hline 
Type &  Schema form & Explanation of examples from Table \ref{tab:categories:2}  \\ 
\hline 
\hline  
1 &   $\text{\textit{pred}}_m \left( m, a \right) $ & Example 1.2: It is enough to know that: 
$\mathcal{S}\left( \text{have} \left( m=[\text{the flower}], a=[\text{pollen}] \right) \right) > \mathcal{S}\left( \text{have} \left( m=[\text{the bee}], a=[\text{pollen}] \right) \right) $  \\ 
\hline    
2 & $ \text{\textit{pred}}_m \left( m, a \right) |  \widehat{pred}_m \left( m, \widehat{a} \right), cn  $ & 
 Example 2.2: It is enough to know that: 
$
  \mathcal{S}\left( \text{\textit{be afraid of}} \left( m=*, a=* \right) |  \text{\textit{get scared}} \left( m=*, \widehat{a}=* \right), \text{because} \right)  >  
$
$
  \mathcal{S}\left( \text{\textit{be afraid of}} \left( a=*, m=* \right) |  \text{\textit{get scared}} \left( m=*, \widehat{a}=* \right), \text{because}  \right) 
$ \\ 
\hline 
\end{tabular}
\caption{Predicate Schemas and examples of the logic behind the schema design. Here $*$ indicates that the argument is dropped, and $\mathcal{S}(.)$ denotes the scoring function defined in the text.}
\label{tab:categories:1}
\end{table*}

In this section we present multiple kinds of knowledge that are needed in order to improve hard coreference problems. 
Table \ref{tab:categories:2} provides two example sentences for each type of knowledge. 
We use $m$ to refer to a mention. 
A mention can either be an entity $e$ or a pronoun $pro$.  
$pred_{m}$ denotes the predicate of $m$ (similarly, $pred_{pro}$ and $pred_{e}$ for pronouns and entities, respectively). For instance, in sentence 1.1 in Table \ref{tab:categories:2}, the predicate of $e_1$ and $e_2$ is $pred_{e_1}=pred_{e_2}=$\text{``perch on"}. 
$cn$ refers to the discourse connective ($cn$=``and" in sentence 1.1).
$a$ denotes an argument of $pred_m$ other than $m$. For example, in sentence 1.1, assuming that $m=e_1$, the corresponding argument is $a=e_2$.  

\begin{table}
\centering
\begin{tabular}{|c|c|}
\hline 
\parbox[t]{2mm}{\multirow{4}{*}{\rotatebox[origin=c]{90}{Type 1}}} 
& $\mathcal{S}\left( \text{\textit{pred}}_m \left( m, a \right) \right)$ \\ 
& $\mathcal{S}\left( \text{\textit{pred}}_m \left( a, m \right) \right)$ \\ 
& $\mathcal{S}\left( \text{\textit{pred}}_m \left( m, * \right) \right)$ \\ 
& $\mathcal{S}\left( \text{\textit{pred}}_m \left( *, m \right) \right)$ \\ 
\hline 
\parbox[t]{2mm}{\multirow{6}{*}{\rotatebox[origin=c]{90}{Type 2}}} 
& $\mathcal{S}\left( \text{\textit{pred}}_m \left( m, a \right) |  \widehat{pred}_m \left( m, \widehat{a} \right), cn   \right)$ \\ 
& $\mathcal{S}\left( \text{\textit{pred}}_m \left( a, m \right) |  \widehat{pred}_m \left( m, \widehat{a} \right), cn   \right)$ \\ 
& $\mathcal{S}\left( \text{\textit{pred}}_m \left( m, a \right) |  \widehat{pred}_m \left( \widehat{a}, m \right), cn   \right)$ \\ 
& $\mathcal{S}\left( \text{\textit{pred}}_m \left( a, m \right) |  \widehat{pred}_m \left( \widehat{a}, m \right), cn   \right)$ \\ 
& $\mathcal{S}\left( \text{\textit{pred}}_m \left( m, * \right) |  \widehat{pred}_m \left( m, * \right), cn   \right)$ \\ 
& { \vdots}\\ 
\hline 
\end{tabular}
\caption{Possible variations for scoring function statistics. Here $*$ indicates that the argument is dropped.}
\label{tab:cat:1:variations}
\end{table}

We represent the knowledge needed with two types of Predicate Schemas (as depicted in Table \ref{tab:categories:1}). 
To solve the assignment of \text{[it]}$_{pro}$ in sentence 1.1, as mentioned in Section \ref{sec:intro}, we need the knowledge that ``a limb bends" is more reasonable than ``a bird bends". 
Note that the predicate of the pronoun is playing a key role here. Also the entity mention itself is essential. 
Similarly, for sentence 1.2, to resolve \text{[it]}$_{pro}$, we need the knowledge that ``bee had pollen" is more reasonable than ``flower had pollen". 
Here, in addition to entity mention and the predicate (of the pronoun), we need the argument which shares the predicate with the pronoun. 
To formally define the type of knowledge needed we denote it with ``$\boldmath{\textbf{\text{\textit{pred}}}_m(m,a)}$" where $m$ and $a$ are a mention and an argument, respectively\footnote{Note that the order of $m$ and $a$ relative to the predicate is a critical issue. To keep things general in the schemas definition, we do not show the ordering; however, when using scores in practice the order between a mention and an argument is a critical issue.}.
We use $\mathcal{S}(.)$ to denote the score representing how likely the combination of the predicate-mention-argument is. 
For each schema, we use several variations by either changing the order of the arguments (\textit{subj.} vs \textit{obj.}) or dropping either of them. We score the various Type 1 and Type 2 schemas (shown in Table \ref{tab:cat:1:variations}) differently. 
%We create variations of the score function by either changing the order of the arguments (\textit{subj.} vs \textit{obj.}) or dropping either of them. The different variations of scores following the Type 1 schema are represented in Table \ref{tab:cat:1:variations} (upper half). 
The first row of Table \ref{tab:categories:1} shows how Type 1 schema is being used in the case of Sentence 1.2.
%we represent the logic for sentence 1.2 again in terms of the score, given the knowledge needed to resolve the pronoun.  

For sentence 2.2, we need to have the knowledge that the \textit{subject} of the verb phrase ``{be afraid of}" is more likely than the \textit{object} of the verb phrase ``{be afraid of}" to be the \textit{subject} of the verb phrase ``{get scared}". The structure here is more complicated than that of Type 1 schema. To make it clearer, we analyze sentence 2.1. In this sentence, the \textit{object} of ``{be robbed by}" is more likely than the \textit{subject} of the verb phrase ``{be robbed by}" to be the \textit{object} of ``{the officer arrest}". We can see in both examples (and for the Type 2 schema in general), that both predicates (the entity predicate and the pronoun predicate) play a crucial role. 
Consequently, we design the Type 2 schema to capture the interaction between the entity predicate and the pronoun predicate. 
In addition to the predicates, we may need mention-argument information. Also, we stress the importance of the discourse connective between entity mention and pronoun; if in either sentence 2.1 or 2.2, we change the discourse connective to ``although", the coreference resolution will completely change. Overall, we can represent the knowledge as ``$\boldmath{\textbf{\text{\textit{pred}}}_m \left( m, a \right) |  \widehat{\textbf{\text{\textit{pred}}}}_m \left( m, \widehat{a} \right), cn}$". Just like for Type 1 schema, we can represent Type 2 schema with a score function for different variations of arguments (lower half of Table \ref{tab:cat:1:variations}). In Table \ref{tab:categories:1}, 
we exhibit this for sentence 2.2. 
%we show the logic for sentence 2.2 with the scoring representation of the schema. 

Type 3 contains the set of instances which cannot be solved using schemas of Type 1 or 2. Two such examples are included in Table \ref{tab:categories:2}. In sentence 3.1 and 3.2, the context containing the necessary information goes beyond our triple representation and therefore this instance cannot be resolved with either of the two schema types. It is important to note that the notion of Predicate Schemas is more general than the Type 1 and Type 2 schemas introduced here. Designing more informative and structured schemas will be essential to resolving additional types of hard coreference instances. 

\section{Constrained ILP Inference} \label{sec:inf}
\textit{Integer Linear Programming} (ILP) based formulations of NLP problems \cite{RothYi04} have been used in a board range of NLP problems and, particularly, in coreference problems \cite{CSRRSR11,DenisBa07}. 
Our formulation is inspired by \newcite{ChangSaRo13}. 
Let $\mathcal{M}$ be the set of all mentions in a given text snippet, and $\mathcal{P}$ the set of all pronouns, such that $ \mathcal{P} \subset \mathcal{M}$. 
We train a coreference model by learning a pairwise mention scoring function. 
Specifically, given a mention-pair $(u,v)\in\mathcal{M}$ ($u$ is the antecedent of $v$), we learn a left-linking scoring function $f_{u,v} = \mathbf{w}^\top \phi(u, v)$, where $\phi(u, v)$ is a pairwise feature vector and $\mathbf{w}$ is the weight vector. We then follow the \textit{Best-Link} approach (Section 2.3 from \newcite{CSRRSR11}) for inference. The ILP problem that we solve is formally defined as follows:
$$
\begin{cases}
\displaystyle
\arg\max_y \sum_{
\substack{ %u, v\\ 
u \in \mathcal{M}, 
v \in \mathcal{M} }
} f_{u,v} y_{u,v} \\ 
\text{s.t.} \, \,\, y_{u,v} \in \lbrace 0, 1\rbrace, \,\,\,\,\, \forall u,v\in \mathcal{M}\\
\,\,\,\,\,\,\,\,\,\, \sum_{
\substack{u<v, u \in \mathcal{M}}	
}  y_{u,v} \leq 1, \,\,\,\,\, \forall v\in \mathcal{M} \\ 
\,\,\,\,\,\,\,\,\,\, \text{\small Constraints from Predicate Schemas Knowledge} \\
\,\,\,\,\,\,\,\,\,\, \text{\small Constraints between pronouns.} 
\end{cases}
$$

Here, $u,v$ are mentions and $y_{u,v}$ is the decision variable to indicate whether or not mention $u$ and mention $v$ are coreferents. As the first constraint shows, $y_{u,v}$ is a binary variable. $y_{u,v}$ equals $1$ if $u,v$ are coreferents and $0$ otherwise. The second constraint indicates that we only choose at most one antecedent to be coreferent with each mention $v$. ($u<v$ represents that $u$ appears beore $v$, thus $u$ is an antecedent of $v$.) In this work, we add constraints from Predicate Schemas Knowledge and between pronouns.

The Predicate Schemas knowledge provides a vector of score values $\mathcal{S}(u,v)$ for mention pairs $\lbrace(u, v) | (u\in\mathcal{M}, v\in\mathcal{P} \rbrace$, which concatenates all the schemas involving $u$ and $v$. Entries in the score vector are designed so that the larger the value is, the more likely $u$ and $v$ are to be coreferents. 
We have two ways to use the score values: 
1) Augumenting the feature vector $\phi(u, v)$ with these scores.
2) Casting the scores as constraints for the coreference resolution ILP in one of the following forms:
\begin{equation}
\small
\begin{cases}
\text{if} \; s_i(u, v) \geq \alpha_{i}s_i(w, v) \Rightarrow y_{u,v} \geq y_{w,v},   \\ 
\text{if} \; s_i(u, v) \geq  s_i(w, v) + \beta_{i} \Rightarrow y_{u,v} \geq y_{w,v}, 
\end{cases}
\label{eq:inequalitiess}
\end{equation}
where $s_i(.)$ is the $i$-th dimension of the score vector $\mathcal{S}(.)$ corresponding to the $i$-th schema represented for a given mention pair. $\alpha_i$ and $\beta_i$ are threshold values which we tune on a development set.\footnote{For the $i_\text{th}$ dimension of the score vector, we choose either $\alpha_i$ or $\beta_i$ as the threshold.} If an inequality holds for all relevant schemas (that is, all the dimensions of the score vector), we add an inequality between the corresponding indicator variables inside the ILP.\footnote{If the constraints dictated by any two dimensions of $\mathcal{S}$ are contradictory, we ignore both of them.} As we increase the value of a threshold, the constraints in (\ref{eq:inequalitiess}) become more conservative, thus it leads to fewer but more reliable constraints added into the ILP.
We tune the threshold values such that their corresponding scores attain high enough accuracy, either in the multiplicative form or the additive form.\footnote{The choice is made based on the performance on the development set.}
Note that, given a pair of mentions and context, we automatically instantiate a collection of relevant schemas, and then generate and evaluate a set of corresponding constraints.
To the best of our knowledge, this is the first work to use such automatic constraint generation and tuning method for coreference resolution with ILP inference. 
In Section \ref{sec:know}, we describe how we acquire the score vectors $\mathcal{S}(u,v)$ for the Predicate Schemas in an unsupervised fashion.

We now briefly explain the pre-processing step required in order to extract the score vector $\mathcal{S}(u,v)$ from a pair of mentions. Define a triple structure $ t_m \triangleq \text{\textit{pred}}_m(m, a_m)$ for any $m \in \mathcal{M}$. 
The subscript $m$ for \textit{pred} and $a$, emphasizes that they are extracted as a function of the mention $m$. 
The extraction of triples is done by utilizing the dependency parse tree from the Easy-first dependency parser \cite{goldberg2010efficient}. We start with a mention $m$, and extract its related predicate and the other argument based on the dependency parse tree and part-of-speech information. To handle multiword predicates and arguments, we use a set of hand-designed rules.
We then get the score vector $\mathcal{S}(u, v)$ by concatenating all scores of the Predicate Schemas given two triples $t_u$, $t_v$. Thus, we can expand the score representation for each type of Predicate Schemas given in Table \ref{tab:categories:1}: 1) For Type 1 schema, $\mathcal{S}(u, v) \equiv \mathcal{S}(pred_v(m=u, a=a_v))$ 
\footnote{In $pred_v(m=u, a=a_v)$ the argument and the predicate are extracted relative to $v$ but the mention $m$ is set to be $u$.}
2) For Type 2 schema, $\mathcal{S}(u, v) \equiv \mathcal{S}(pred_u(m=u, a=a_u)|\widehat{pred}_v(m=v, a=a_v),cn)$. 

In additional to schema-driven constraints, we also apply constraints between pairs of pronouns within a fixed distance\footnote{We set the distance to be 3 sentences.}. For two pronouns that are semantically different (e.g. \textit{he} vs. \textit{it}), they must refer to different antecedents. For two non-possesive pronouns that are related to the same predicate (e.g. \textit{\underline{he} saw \underline{him}}), they must refer to different antecedents.\footnote{Three cases are considered: \textit{he}-\textit{him}, \textit{she}-\textit{her}, \textit{they}-\textit{them}} 

\section{Knowledge Acquisition} \label{sec:know}
\begin{table*}
\centering
\small
$\mathcal{S}_{pol}(u, v) = 
\begin{bmatrix}
\mathbf{1} \lbrace \text{\textit{Po}}(p_u) = + \; \text{ AND } \; \text{\textit{Po}}(p_v)=+ \rbrace 
\quad  \text{ OR }\; \quad \; \mathbf{1} \lbrace \text{\textit{Po}}(p_u) = -  \text{ AND } \text{\textit{Po}}(p_v)=- \rbrace \\
\mathbf{1} \lbrace \text{\textit{Po}}(p_u) = + \; \text{ AND } \; \text{\textit{Po}}(p_v)=+ \rbrace \\
\mathbf{1} \lbrace \text{\textit{Po}}(p_u) = -\;  \text{ AND } \; \text{\textit{Po}}(p_v)=- \rbrace \\
\end{bmatrix}
$
\caption{Extrating the polarity score given polarity information of a mention-pair $(u,v)$. To be brief, we use the shorthand notation $p_v \triangleq pred_v$ and $p_u \triangleq pred_u$. $\mathbf{1}\{\cdot\}$ is an indicator function. $s_{pol}(u, v)$ is a binary vector of size three.}
\label{tab:polarity}
\end{table*}

One key point that remains to be explained is how to acquire the knowledge scores $\mathcal{S}(u,v)$. In this section, we propose multiple ways to acquire these scores. 
In the current implementation, we make use of four resources. Each of them generates its own score vector.
Therefore, the overall score vector is the concatenation of the score vector from each resource:  $\mathcal{S}(u,v) = [\mathcal{S}_{giga}(u,v)\; \mathcal{S}_{wiki}(u,v)\;  \mathcal{S}_{web}(u,v)\;  \mathcal{S}_{pol}(u,v) ]$.

\subsection{Gigaword Co-occurence}
We extract triples $t_m \triangleq \text{\textit{pred}}_m(m, a_m)$ (explained in Section \ref{sec:inf}) from Gigaword data (4,111,240 documents). 
We start by extracting noun phrases using the Illinois-Chunker \cite{PunyakanokRo01}. For each noun phrase, we extract its head noun and then extract the associated predicate and argument to form a triple.

We gather the statistics for both schema types after applying lemmatization on the predicates and arguments. Using the extracted triples, we get a score vector from each schema type: $\mathcal{S}_{giga} = [\mathcal{S}^{(1)}_{giga} \; \mathcal{S}^{(2)}_{giga} ]$. 

To extract scores for Type 1 Predicate Schemas, we create occurence counts for each schema instance. After all scores are gathered, our goal is to query $\mathcal{S}^{(1)}_{giga}(u, v) \equiv \mathcal{S}(pred_v(m=u, a=a_v))$ from our knowledge base. The returned score is the $log(.)$ of the number of occurences. 

For Type 2 Predicate Schemas, we gather the statistics of triple co-occurence.  
We count the co-occurrence of neighboring triples that share at least one linked argument. We consider two triples to be neighbors if they are within a distance of three sentences. We use two heuristic rules to decide whether a pair of arguments between two neighboring triples are coreferents or not: 1) If the head noun of two arguments can match, we consider them coreferents. 2) If one argument in the first triple is a person name and there is a compatible pronoun (based on its gender and plurality information) in the second triple, they are also labeled as coreferents. We also extract the discourse connectives between triples (\textit{because}, \textit{therefore}, etc.) if there are any.
To avoid sparsity, we only keep the mention roles (only \textit{subj} or \textit{obj}; no exact strings are kept). Two triple-pairs are considered different if they have different predicates, different roles, different coreferred argument-pairs, or different discourse connectives. The co-occurrence counts extracted in this form correspond to Type 2 schemas in Table \ref{tab:categories:1}. During inference, we match a Type 2 schema for $ \mathcal{S}^{(2)}_{giga}(u, v) \equiv \mathcal{S}(pred_u(m=u, a=a_u)|\widehat{pred}_v(m=u, a=a_v), cn) $.

Our method is related, but different from the proposal in \newcite{balasubramanian2012rel}, who suggested to extract triples using an OpenIE system \cite{ollie-emnlp12}. We extracted triples by starting from a mention, then extract the predicate and the other argument. An OpenIE system does not easily provide this ability. Our Gigaword counts are gathered also in a way similar to what has been proposed in \newcite{chambers2009unsupervised}, but we gather much larger amounts of data.

\subsection{Wikipedia Disambiguated Co-occurence}
One of the problems with blindly extracting triple counts is that we may miss important semantic information. To address this issue, we use the publicly avaiable Illinois Wikifier \cite{ChengRo13,ratinov-EtAl:2011:ACL-HLT2011}, a system that disambiguates mentions by mapping them into correct Wikipedia pages, to process the Wikipedia data. We then extract from the Wikipedia text all entities, verbs and nouns, and gather co-occurrence statistics with these syntactic \textit{variations}: 1) \textit{immediately after} 2) \textit{immediately before}  3) \textit{before} 4) \textit{after}. For each of these variations, we get the probability and count\footnote{We use the $log(.)$ of the counts here.} of a pair of words (e.g. probability\footnote{Conditional probability of ``limb" immediately following the given verb ``bend".}/count for ``bend" \textit{immediately following} ``limb") as separate dimensions of the score vector.

Given the co-occurrence information, we get a score vector $\mathcal{S}_{wiki}(u, v)$ corresponding to Type 1 Predicate Schemas, and hence $\mathcal{S}(u, v)_{wiki} \equiv \mathcal{S}(pred_v(m=u, a=a_v))$. 

\subsection{Web Search Query Count}
Our third source of score vectors is web queries that we implement using Google queries. We extract a score vector $\mathcal{S}_{web}(u, v) \equiv \mathcal{S}(pred_v(m=u, a=a_v))$ (Type 1 Predicate Schemas) by querying for 1) ``$u$ $a_v$'' 2) ``$u$ $pred_v$'' 3) ``$u$ $pred_v$ $a_v$'' 4) ``$a_v$ $u$''\footnote{We query this only when $a_v$ is an adjective and $pred_v$ is a to-be verb.}. For each variation of nouns (plural and singular) and verbs (different tenses) we create a different query and average the counts over all queries. Concatenating the counts (each is a separate dimension) would give us the score vector $\mathcal{S}_{web}(u, v)$.

\begin{table*} %[htbp]
\centering
%\footnotesize
\begin{tabular}{|l|c|c|c|c|c|c|}
\hline
		   & \# Doc & \# Train & \# Test & \# Mention & \# Pronoun & \# Predictions for Pronoun \\
\hline  
Winograd   & 1886 & 1212   & 674   & 5658     & 1886     & 1348 					\\
\hline
WinoCoref  & 1886 & 1212   & 674   & 6404     & 2595     & 2118						\\
\hline
ACE   & 375  & 268    & 107   & 23247    & 3862     & 13836					\\
\hline
OntoNotes & 3150 & 2802   & 348   & 175324   & 58952    & 37846					\\
\hline
\end{tabular}
\caption{Statistics of \textit{Winograd}, \textit{WinoCoref}, \textit{ACE} and \textit{OntoNotes}. We give the total number of mentions and pronouns, while the number of predictions for pronoun is specific for the test data. We added 746 mentions (709 among them are pronouns) to \textit{WinoCoref} compared to \textit{Winograd}.} \label{tab:dataset}
\end{table*}

\subsection{Polarity of Context}
Another rich source of information is the polarity of context, which has been previously used for Winograd schema problems \cite{rahman2012resolving}. Here we use a slightly modified version. 
The polarity scores are used for Type 1 Predicate Schemas and therefore we want to get $\mathcal{S}_{pol}(u, v) \equiv \mathcal{S}(pred_v(m=u, a=a_v))$. 
We first extract polarity values for \textit{Po}($pred_u$) and \textit{Po}($pred_v$) by repeating the following procedures for each of them: 
\begin{itemize}
\setlength{\itemsep}{0.01pt}
 \setlength{\parskip}{0pt}
 \setlength{\parsep}{0pt}
\item We extract initial polarity information given the predicate (using the data provided by \newcite{wilson2005opinionfinder}). 
\item If the role of the mention is \textit{object}, we negate its polarity. 
\item If there is a polarity-reversing discourse connective (such as ``but") preceding the predicate, we reverse the polarity. 
\item If there is a negative comparative adverb (such as ``less", ``lower") we reverse the polarity.
\end{itemize}
Given the polarity values \textit{Po}($pred_u$) and \textit{Po}($pred_v$), we construct the score vector $\mathcal{S}_{pol}(u, v)$ following Table \ref{tab:polarity}.  

\section{Experiments} \label{sec:exp}
In this section, we evaluate our system for both hard coreference problems and general coreference problems, and provide detailed anaylsis on the impact of our proposed Predicate Schemas. Since we treat resolving hard pronouns as part of the general coreference problems, we extend the \textit{Winograd} dataset with a more complete annotation to get a new dataset. We evaluate our system on both datasets, and show significant improvemnt over the baseline system and over the results  reported in \newcite{rahman2012resolving}. Moreover, we show that, at the same time, our system achieves the state-of-art performance on standard coreference datasets.

\subsection{Experimental Setup} \label{sec:expset}
\noindent {\bf Datasets:} Since we aim to solve hard coreference problems, we choose to test our system on the \textit{Winograd} dataset\footnote{Available at http://www.hlt.utdallas.edu/\~{}vince/data/emnlp12/} \cite{rahman2012resolving}.
It is a challenging pronoun resolution dataset which consists of sentence pairs based on Winograd schemas. The original annotation only specifies one pronoun and two entites in each sentence, and it is considered as a binary decision for each pronoun. As our target is to model and solve them as general coreference problems, we expand the annotation to include all pronouns and their linked entities as mentions (We call this new re-annotated dataset \textit{WinoCoref}\footnote{Available at http://cogcomp.cs.illinois.edu/page/data/}).
Ex.3 in Section \ref{sec:intro} is from the \textit{Winograd} dataset. It originally only specifies \textit{he} as the pronoun in question, and we added \textit{him} and \textit{his} as additional target pronouns. We also use two standard coreference resolution datasets \textit{ACE}(2004) \cite{NIST04} and \textit{OntoNotes-5.0} \cite{PRMPWX11} for evaluation. Statistics of the datasets are provided in Table \ref{tab:dataset}.

\noindent {\bf Baseline Systems:} We use the state-of-art Illinois coreference system as our baseline system \cite{ChangSaRo13}. It includes two different versions. One employs \textit{Best-Left-Link} (BLL) inference method \cite{NgCa02}, and we name it \textit{Illinois}\footnote{In implementation, we use the L$^3$M model proposed in \newcite{ChangSaRo13}, which is slightly different. It can be seen as an extension of BLL inference method.}; while the other uses ILP with constraints for inference, and we name it \textit{IlliCons}. Both systems use \textit{Best-Link Mention-Pair} (BLMP) model for training. On \textit{Winograd} dataset, we also treat the reported result from \newcite{rahman2012resolving} as a baseline.

\noindent {\bf Developed Systems:} We present three variations of the Predicate Schemas based system developed here. We inject Predicate Schemas knowledge as mention-pair features and retrain the system (\textit{KnowFeat}). We use the original coreference model and Predicate Schemas knowledge as constraints during inference (\textit{KnowCons}). We also have a combined system (\textit{KnowComb}), which uses the schema knowledge to add features for learning as well as constraints for inference. A summary of all systems is provided in Table \ref{tab:system}.

\begin{table} %[htbp]
\footnotesize
\centering
\begin{tabular}{|c|c|c|}
\hline 
Systems & Learning Method & Inference Method \\
\hline 
Illinois & BLMP & BLL \\
\hline 
IlliCons & BLMP & ILP \\
\hline 
KnowFeat & BLMP+SF & BLL \\
\hline 
KnowCons & BLMP & ILP+SC \\
\hline 
KnowComb & BLMP+SF & ILP+SC \\
\hline 
\end{tabular}
\caption{Summary of learning and inference methods for all systems. SF stands for schema features while SC represents constraints from schema knowledge.}
\label{tab:system}
\end{table}

\begin{table*} %[htbp]
\centering
\footnotesize
\begin{tabular}{|l|c|c|c|c|c|c|c|c|}
\hline
Dataset & Metric	 &	Illinois & IlliCons & \newcite{rahman2012resolving} &  KnowFeat & KnowCons & KnowComb \\
\hline
\textit{Winograd} & Precision &	51.48 & 53.26 & 73.05 & 71.81 & 74.93 & \textbf{76.41} \\
\hline	
\textit{WinoCoref} & AntePre  & 68.37 & 74.32 & ----- & 88.48 & 88.95 & \textbf{89.32} \\ 
\hline
\end{tabular}
\caption{Performance results on \textit{Winograd} and \textit{WinoCoref} datasets. All our three systems are trained on \textit{WinoCoref}, and we evaluate the predictions on both datasets. Our systems improve over the baselines by over than 20\% on \textit{Winograd} and over 15\% on \textit{WinoCoref}.} 
\label{tab:Wino}
\end{table*}

\noindent {\bf Evaluation Metrics:} When evaluating on the full datasets of \textit{ACE} and \textit{OntoNotes}, we use the widely recoginzed metrics MUC \cite{VBACH95}, BCUB \cite{BaggaBa98}, Entity-based CEAF (CEAF$_e$) \cite{Luo05} and their average. As \textit{Winograd} is a pronoun resolution dataset, we use precision as the evaluation metric. Although \textit{WinoCoref} is more general, each coreferent cluster only contains 2-4 mentions and all are within the same sentence. Since traditional coreference metrics cannot serve as good metrics, we extend the precision metric and design a new one called \textit{AntePre}. Suppose there are $k$ pronouns in the dataset, and each pronoun has $n_1,n_2,\cdots,n_k$ antecedents, respectively. We can view predicted coreference clusters as binary decisions on each antecedent-pronoun pair (linked or not). The total number of binary decisions is $\sum_{i=1}^k n_i$. We then meaure how many binary decisions among them are correct; let $m$ be the number of correct decisions, then \textit{AntePre} is computed as: 
$\frac{m}{\sum_{i=1}^k n_i}$.

\subsection{Results for Hard Coreference Problems}
Performance results on \textit{Winograd} and \textit{WinoCoref} datasets are shown in Table \ref{tab:Wino}. The best performing system is \textit{KnowComb}. It improves by over 20\% over a state-of-art general coreference system on \textit{Winograd} and also outperforms \newcite{rahman2012resolving} by a margin of 3.3\%. On the \textit{WinoCoref} dataset, it improves by 15\%. These results show significant performance improvement by using Predicate Schemas knowledge on hard coreference problems. Note that the system developed in \newcite{rahman2012resolving} cannot be used on the \textit{WinoCoref} dataset. The results also show that it is better to compile knowledge into constraints when the knowledge quality is high than add them as features.

\subsection{Results for Standard Coreference Problems}
Performance results on standard \textit{ACE} and \textit{OntoNotes} datasets are shown in Table \ref{tab:Standard}. Our \textit{KnowComb} system achieves the same level of performance as does the state-of-art general coreference system we base it on. As hard coreference problems are rare in standard coreference datasets, we do not have significant performance improvement. However, these results show that our additional Predicate Schemas do not harm the predictions for regular mentions.

\begin{table} %[htb]
%\scriptsize
\centering
\begin{tabular}{|l|c|c|c|c|}
\hline
System	 &	MUC   &	BCUB  &	CEAFe &	AVG	\\
\hline
\multicolumn{5}{|c|}{ACE} \\
\hline
IlliCons &	\textbf{78.17} &	81.64 &	\textbf{78.45} & \textbf{79.42}	\\
\hline
KnowComb &	77.51 &	\textbf{81.97} &	77.44 &	78.97	\\
\hline
\multicolumn{5}{|c|}{OntoNotes} \\
\hline
IlliCons &	84.10 &	\textbf{78.30} &	\textbf{68.74} & \textbf{77.05}	\\
\hline
KnowComb &	\textbf{84.33} &	78.02 &	67.95 &	76.76	\\
\hline
\end{tabular}
\caption{Performance results on \textit{ACE} and \textit{OntoNotes} datasets. Our system gets the same level of performance compared to a state-of-art general coreference system.} 
\label{tab:Standard}
\end{table}

\subsection{Detailed Analysis}
To study the coverage of our Predicate Schemas knowledge, we label the instances in \textit{Winograd} (which also applies to \textit{WinoCoref}) with the type of Predicate Schemas knowledge required. The distribution of the instances is shown in Table \ref{tab:distribution:in:categiries}. Our proposed Predicate Schemas cover 73\% of the instances.

\begin{table} %[ht]
\centering
%\footnotesize
\begin{tabular}{|l|c|c|c|}
\hline 
Category & Cat1 & Cat2 & Cat3 \\ 
\hline 
Size & 317 & 1060 & 509 \\ 
\hline 
Portion & 16.8\% & 56.2\% & 27.0\% \\ 
\hline 
\end{tabular}
\caption{Distribution of instances in \textit{Winograd} dataset of each category. Cat1/Cat2 is the subset of instances that require Type 1/Type 2 schema knowledge, respectively. All other instances are put into Cat3. Cat1 and Cat2 instances can be covered by our proposed Predicate Schemas.}
\label{tab:distribution:in:categiries}
\end{table}

We also provide an ablation study on the \textit{WinoCoref} dataset in Table \ref{tab:expAblationSource}. These results use the best performing \textit{KnowComb} system. They show that both Type 1 and Type 2 schema knowledge have higher precision on Category 1 and Category 2 data instances, respectively, compared to that on full data. Type 1 and Type 2 knowledge have similiar performance on full data, but the results show that it is harder to solve instances in category 2 than those in category 1. Also, the performance drop between Cat1/Cat2 and full data indicates that there is a need to design more complicated knowledge schemas and to refine the knowledge acquisition for further performance improvement. 

\begin{table} %[ht]
\centering
%\footnotesize
\begin{tabular}{|l|c|c|}
\hline
Schema   & AntePre(Test)	& AntePre(Train)	\\
\hline
Type 1   & 76.67	        & 86.79	\\
\hline
Type 2   & 79.55	        & 88.86 \\
\hline
Type 1 (Cat1) & 90.26	&  93.64	\\
\hline
Type 2 (Cat2) & 83.38	&  92.49	\\
\hline
\end{tabular}
\caption{Ablation Study of Knowledge Schemas on \textit{WinoCoref}. The first line specifies the preformance for \textit{KnowComb} with only Type 1 schema knowledge tested on all data while the third line specifies the preformance using the same model but tested on Cat1 data. The second line specifies the preformance results for \textit{KnowComb} system with only Type 2 schema knowledge on all data while the fourth line specifies the preformance using the same model but tested on Cat2 data.} \label{tab:expAblationSource}
\end{table}

\section{Related Work} \label{sec:rel}
\noindent  {\bf{Winograd Schema:}}  \newcite{winograd1972understanding} showed that small changes in context could completely change coreference decisions. \newcite{levesque2011winograd} proposed to assemble a set of sentences which comply with Winograd's schema. 
Specifically, there are pairs of sentences which are identical except for minor differences which lead to different references of the same pronoun in both sentences.  
These references can be easily solved by humans, but are hard, he claimed, for computer programs.

\noindent {\bf{Anaphora Resolution:}}
There has been a lot of work on anaphora resolution in the past two decades. Many of the early rule-based systems like \newcite{hobbs1978resolving} and \newcite{lappin1994algorithm} gained considerable popularity. The early designs were easy to understand and the rules were designed manually. 
With the development of machine learning based models \cite{ConnollyEtAl:94,soon2001machine,ng2002identifying}, attention shifted to solving standard coreference resolution problems. However, many hard coreference problems involve pronouns. As Winograd's schema shows, there is still a need for further investigation in this subarea.

\noindent {\bf{World Knowledge Acquisition:}}
Many tasks in NLP (such as Textual Entailment, Question Answering, etc.) require \textit{World Knowledge}. 
Although there are many existing works on acquiring them \cite{schwartz2009acquiring,balasubramanian2013generating,TandonDeMeloWeikum2014}, there is still no consensus on how to represent, gather and utilize high quality \textit{World Knowledge}. When it comes to coreference resolution, there are a handful of works which either use web query information or apply alignment to an external knowledge base \cite{RahmanNg11c,kobdani2011bootstrapping,RatinovRo12,BansalKl12,zheng13:dynamic}. With the introduction of Predicate Schema, our goal is to bring these different approaches together and provide a coherent view.

\section*{Acknowledgments}
The authors would like to thank Kai-Wei Chang, Alice Lai, Eric Horn and Stephen Mayhew for comments that helped to improve this work.
This work is partly supported by NSF grant \#SMA 12-09359 and by DARPA under agreement number FA8750-13-2-0008. The U.S. Government is authorized to reproduce and distribute reprints for Governmental purposes notwithstanding any copyright notation thereon. The views and conclusions contained herein are those of the authors and should not be interpreted as necessarily representing the official policies or endorsements, either expressed or implied, of DARPA or the U.S. Government. 

\bibliography{new,cited-long20121212,ccg-long20121212}
\end{document}